\documentclass[journal]{IEEEtran}
\usepackage{amssymb,amsmath,bm}
\usepackage{mathrsfs}
\usepackage[lofdepth,lotdepth]{subfig}
\usepackage{booktabs}

\usepackage{mwe}

\usepackage{framed,multirow}
\usepackage{graphicx}

\usepackage{amssymb}
\usepackage{latexsym}

\usepackage{url}
\usepackage{times}
\usepackage{xcolor}
\usepackage{times}
\usepackage{epsfig}
\usepackage{graphicx}
\usepackage{amsmath}
\usepackage{amssymb}
\usepackage{algorithm,algorithmic}
\usepackage{url}
\usepackage{array}
\usepackage{bm}
\usepackage{verbatim}
\usepackage{placeins}
\usepackage{multicol}
\usepackage{lipsum}
\usepackage{tikz}
\usepackage{caption}
\usepackage{subfig}
\usepackage{rotating}

\usepackage{multirow}

\newlength{\tempheight}
\newlength{\tempwidth}

\newcommand{\rowname}[1]
{\rotatebox{90}{\makebox[\tempheight][c]{\textbf{#1}}}}

\newcommand{\columnname}[1]
{\makebox[\tempwidth][c]{\textbf{#1}}}

\graphicspath{ {Images/} }

\ifCLASSINFOpdf
\else
\fi

\begin{document}
%
\title{Two-stream Encoder-Decoder Network for Localizing Image Forgeries}
%
%
%

\author{\IEEEauthorblockN{Aniruddha Mazumdar  and 
        Prabin Kumar Bora}\\ 
\thanks{Aniruddha Mazumdar and Prabin Kumar Bora are with the Department
of Electronics and Electrical Engineering, Indian Institute of Technology Guwahati, Assam-781039, India e-mail: (m.aniruddha@iitg.ac.in; prabin@iitg.ac.in).}
}

\maketitle

\begin{abstract}
 This paper proposes a novel two-stream encoder-decoder network, which utilizes both the high-level and the low-level image features for precisely localizing forged regions in a manipulated image. This is motivated from the fact that the forgery creation process generally introduces both the high-level artefacts (e.g. unnatural contrast) and the low-level artefacts (e.g. noise incosistency) to the forged images. In the proposed two-stream network, one stream learns the low-level manipulation-related features in the encoder side by extracting noise residuals through a set of high-pass filters in the first layer of the encoder network. In the second stream, the encoder learns the high-level image manipulation features from the input image RGB values. The coarse feature maps of both the encoders are upsampled by their correspondin decoder network to produce dense feature maps. The dense feature maps of the two streams are concatenated and fed to a final convolutional layer with sigmoidal activation to produce pixel-wise prediction. We have carried out experimental analysis on multiple standard forensics datasets to evaluate the performance of the proposed method. The experimental results show the efficacy of the proposed method with respect to the state-of-the-art.
\end{abstract}


%
\IEEEpeerreviewmaketitle

\section{Introduction}\label{sec:intro}

Nowadays, the creation of visually plausible forged images has become a simple task due to various sophisticated image editing software like Adobe Photoshop and GIMP. Sometimes these realistic-looking forged images are used with malicious intentions. Image forensics aims to develop techniques that can differentiate the forged images from the authentic ones. There are different ways to forge an image; \emph{splicing}, \emph{copy-move}, and \emph{content-removal} are the main types. In splicing, contents of different images are copied and pasted onto a single image to create a composite one. In copy-move forgery, contents from a single image are cloned at different locations of the same image itself. In content removal forgery, regions from an image are removed and filled with inpainting pixels to hide some information from the image.

A perfectly curated forged image looks visually plausible; Nevertheless, the forgery creation process introduces various \emph{high-level} and \emph{low-level} artefacts in the forged image. The high-level artefacts include unnatural contrast and edges, inconsistent out-of-focus blur, etc.  The low-level artefacts are inconsistent noise levels in different parts of a forged image, traces related to double JPEG-compression, and different types of post-processing operations carried out on the forged regions etc. Different image forensics methods utilize these high-level and low-level artefacts to expose various types of forgeries.


Most of the previous forensics methods have focused on detecting a single type of forgery, e.g., detecting splicing or copy-move forgeries. These works are based on some assumptions about the type of forgery. The methods developed for detecting splicing assume that the spliced parts will have certain features (i.e., high-level or low-level artefacts) that will not match with that of the authentic parts. For example, in \cite{carvalho_2} illumination-based high-level features and in \cite{Chen_PRNU} camera sensor-based low-level features  are utilized to expose splicing forgery. The methods aimed at detecting copy-move forgeries assume the presence of cloned objects in the forged image. For example, Popescu and Farid \cite{popescu} proposed to use a high-level image feature to find the duplicate regions present in a forged image.

Although the above-mentioned methods are effective in detecting a single type forgery, they may not be able to detect other forgeries. This is because the methods developed to detect a particular type of forgery make some assumptions about the forgery. To tackle this, a number of methods \cite{zhou}, \cite{bappy}, \cite{mantra}, \cite{mag} have been proposed recently, which can detect multiple forgeries, such as splicing, copy-move and content-removal, in a single framework. These methods are based on learning different high-level and low-level features using different deep learning-based techniques. 


This paper proposes a novel two-stream encoder-decoder network to detect and localize multiple forgeries in a single framework. One stream of the encoder learns to differentiate between the authentic and forged parts based on the high-level image features. The other stream learns the inconsistencies in the low-level image features between the authentic and forged regions present in a forged images. The main highlights of the paper are as follows: We have developed a two-stream encoder-decoder network that learns both the high-level and the low-level manipulation-related features for pixel-wise forgery localization. We have explored different two-stream architectures and also experimented with different loss functions. The network is trained end-to-end and hence can be fine-tuned for any type of forgeries.

The rest of the paper is organized as follows: Section 2 describes the existing forensics methods for detecting different types of forgeries. Section 3 presents the proposed forensics method. Section 4 discusses the experimental results, and Section 5 concludes the paper.

\section{Related Work and Motivation}
Many methods have been proposed in the image forensics literature for the detection of different types of forgeries. The earlier approaches were aimed at detecting or localizing a specific type of forgery, e.g. detecting splicing or copy-move forgery. These methods were based on the detection of forgery-specific traces present in the manipulated images. 

For the detection of splicing forgeries, traces such as the differences in the local noise-levels \cite{NOI2} and the JPEG compression levels \cite{ADQ1}, \cite{BLK} at different locations in an image, and the mismatch in the colour filter array (CFA) interpolation methods \cite{CFA2} are used as hand-crafted features. Recently, deep learning-based methods are proposed, which learn the splicing-related traces by training convolutional neural networks (CNNs) \cite{bondi2017tampering}, \cite{tiwari2019detection}, \cite{mfcn} and siamese networks \cite{cozzolino2019noiseprint}. For example, Salloum \emph{et al.} \cite{mfcn} proposed a multi-task fully-convolutional network (MFCN) to localize the forged regions present in a manipulated image. Cozzolino and Verdoliva \cite{cozzolino2019noiseprint} proposed to localize splicing forgeries by learning camera-related traces, through training a siamese network to differentiate between image patches coming from the same and different images.

For the detection of copy-move forgery, visual features are computed from different parts of an image and checked for the presence of duplicate regions \cite{popescu}. Ardizzone \emph{et al.} \cite{ardizzone2015copy} proposed to detect cloned regions by matching triangles of keypoints. Wu \emph{et al.} \cite{wu2018busternet} proposed a two-branch CNN, where one branch learns to check for cloned objects based on visual similarities, and the other branch learns to check for manipulated regions based on visual artifacts.

Although the above methods can detect/localize splicing and copy-move forgeries effectively, they have the limitation of being able to detect a single type of forgery only. However, in real forensics scenarios, the type of forgery is generally unknown beforehand. There might be more than one type of forgeries present in a single manipulated image. These concerns necessitate the development of forensics techniques that can detect different types of forgeries in a single framework.  

The image forensics community is currently focusing on developing deep learning-based methods to detect and localize multiple forgeries in a single framework. Zhou \emph{et al.} \cite{zhou} proposed a two-stream faster R-CNN network (RGB-N) for localizing different forgeries. One stream of the network learns the high-level manipulation-related features, while the other stream learns the low-level manipulation-related features. Bappy \emph{et al.} \cite{bappy} proposed a two-stream method (LSTM-EnDec), where one stream applies an LSTM network on the hand-crafted resampling features to produce final low-level features, and the other stream learns high-level features using a CNN-based encoder network. The resampling features, computed using the Radon transform and the Laplacian filters, help detect different types of operations carried out on the forged regions while creating a forgery. The encoder network learns various high-level manipulation-related traces, such as unnatural contrast. Finally, the features of both streams are concatenated, and then a single decoder network is applied to produce a pixel-wise prediction map. Wu \emph{et al.} \cite{mantra} proposed another deep learning-based method (ManTra Net) that first extracts image manipulation trace-related features from a test image and then checks the consistency between them to localize the forged regions. Kniaz \emph{et al.} \cite{mag} have proposed to train a discriminative segmentation model in a mixed adversarial setting using a GAN (MAG) to localize different types of forgeries.

The state-of-the-art methods \cite{zhou}, \cite{bappy} have shown that the fusion of high-level and low-level manipulation-related traces helps in detecting and localizing different forgeries more effectively. For example, Zhou \emph{et al.} \cite{zhou} showed the effectiveness of the fusion of the high-level and the low-level features, learned from the training examples, in an R-CNN framework. However, the method can only give the bounding box-level localization of the forged regions, and hence not a true pixel-wise localization. The bounding box-level localization may include many authentic pixels inside the forged bounding box depending upon the shape of the forged regions. The LSTM-EnDec fused the low-level features, computed using Radon transform and LSTM network, and the high-level features, computed using the CNN-based encoder network, for pixel-wise localization of forged regions. However, as the low-level manipulation traces are hand-crafted features, they may not be optimal for the forgery localization task. Based on these motivations, we propose to employ a two-stream encoder-decoder network which learns both the low-level and high-level manipulation-related traces automatically from the training images and can perform pixel-wise forgery localization.


\section{Proposed Method}

The proposed method aims at localizing forgeries present in a manipulated image though a two-stream encoder-decoder neural network. One stream of the encoder-decoder network is the \emph{image-stream}, which learns the high-level manipulation traces present in forged images. The other stream is the \emph{noise-stream}, which learns the low-level manipulation traces from the noise residuals computed from the input image. The motivation for employing a two-stream network comes from the nature of artefacts present in a forged image. When an image is manipulated to create forgeries, such as splicing, different types of artefacts are introduced in the forged image. These artefacts can be broadly divided into the following two categories:
\begin{enumerate}
	\item High-level artefacts: The high-level artefacts, generally introduced in a forged image, include artificial edges, unnatural contrasts, inconsistent blur, etc. For example, the forged regions in a splicing or a copy-move forgery may have slightly high contrast than the rest of the image. When an object is copied and pasted on a different location of an image, the edges around the pasted object tend be different from the natural object edges.
	
	\item Low-level artefacts: The low-level artefacts, present in forged images, are the inconsistencies in the noise levels between the forged and the authentic regions. For example, in a spliced image, the spliced regions may come from different images, and possibly with different noise levels from the rest of the image. Also, the application of various image editing operations, such as Gaussian blurring, contrast enhancement, on an image change the local dependencies between the neighbouring pixels in unique ways. This, in turn, modifies the noise residuals present in the images distinctively for different editing operations. In a forged image, the forged regions are generally edited using different post-processing operations to make it look visually plausible. Therefore, the application of different editing operations on a forged image also contributes to the low-level artefacts.
\end{enumerate}

The proposed two-stream encoder-decoder network learns the features related to both the high-level and the low-level manipulation traces in the encoder side and upsamples these features to produce dense feature maps in the decoder side. The dense feature maps from both the streams are concatenated and fed to another convolution layer to produce the output prediction map, representing the forged and authentic pixels.


\begin{figure*}[]
	\centering
	\includegraphics[width=18.0cm, height=9cm]{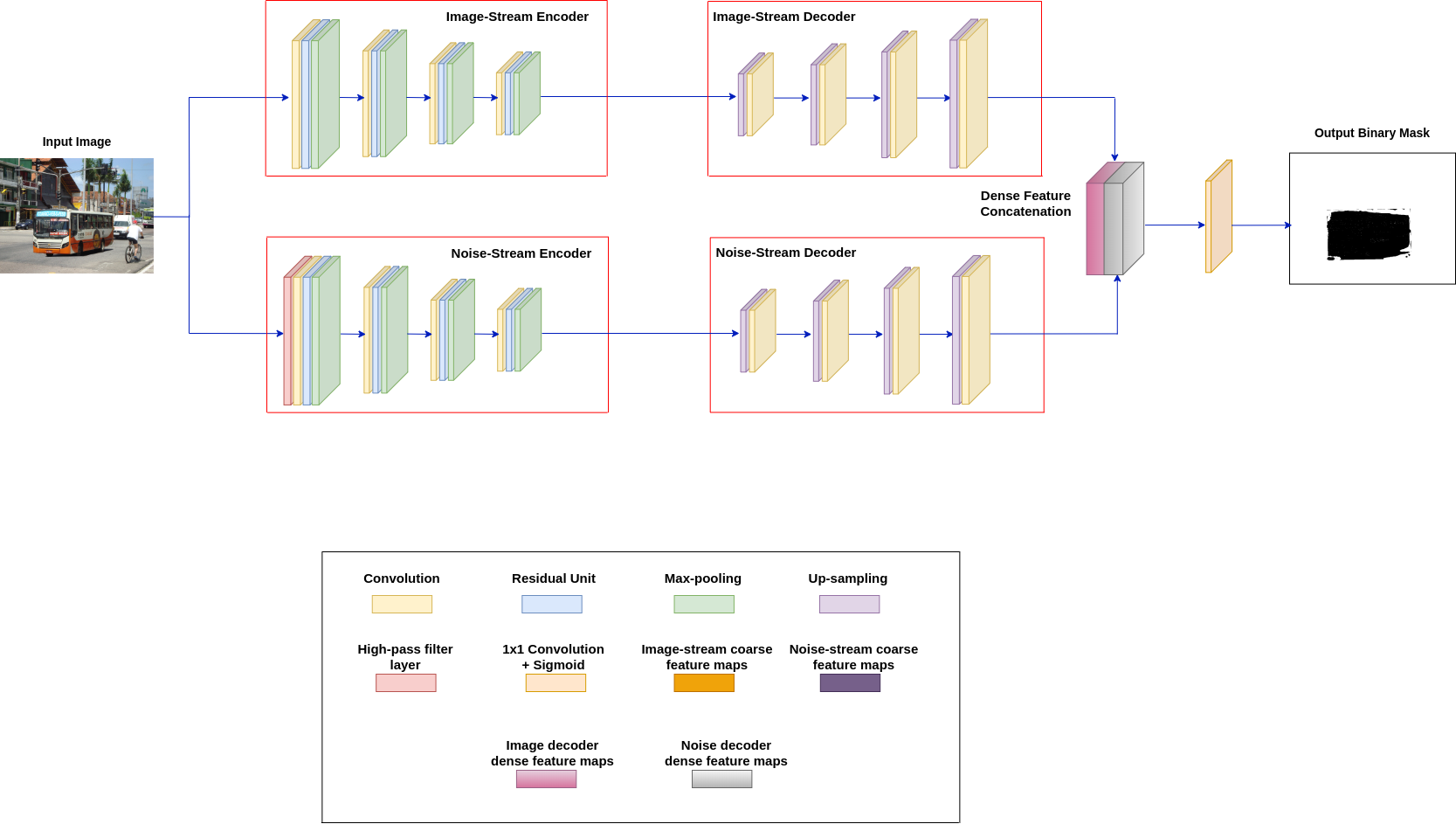} 
	
	\caption{Block diagram of the proposed two-stream encoder-decoder network. The encoder in the Image-stream learns high-level manipulation traces, such as artificial contrast. The encoder in the noise-stream learns the low-level traces, such as noise inconsistency, by employing a high-pass filter layer at the begining of the network. The decoders in both the streams upsamples the coarse feature maps of the encoders to produce dense feature maps, which are then concatenated and fed to using a $1\times1$ sigmoidal convolutional layer for performing the pixel-wise classification.} \label{EnDec_block}
\end{figure*}

Figure \ref{EnDec_block} shows the block diagram of the proposed network. As shown in the figure, there are two encoder-decoder streams in the proposed network, namely the image-stream encoder-decoder (ISED) and the noise-stream encoder-decoder (NSED). Given a test image as input, it is processed by both the ISED and the NSED. The encoder in each stream performs a series of convolution, non-linear mapping, and max-pooling operations on the input image and produces the coarse feature maps. The coarse feature maps capture the relevant information about the high-level and the low-level manipulation-related traces. The corresponding decoder for each stream then performs upsampling and convolution operations on the coarse feature maps to produce the dense feature maps. The dense feature maps are of the same resolution as the input image, and helpful for pixel-wise classification. The output dense feature maps of both stream decoders are concatenated and fed to $1\times1$ sigmoidal convolutional layer to produce the pixel-wise probability map. The pixel-wise probability map gives the probability of each pixel belonging to the authentic and the forged classes.

Both the streams of the encoder-decoder network are described below.


%

\subsection{Image-Stream Encoder-Decoder (ISED)}
The aim of this stream is to learn the high-level manipulation traces that are left behind by different forgeries. The encoder of the ISED is a fully-convolutional neural network (FCN) operating on the RGB values of the input image. It consists of $4$ convolutional layers, $4$ residual blocks and $4$ max-pooling layers. At each convolutional layer, the rectified linear unit (ReLU) activation and the batch normalization technique are used. Each residual block has $3$ convolutional layers along with the ReLU activation and the batch normalization applied after each convolution operation. The output of each residual block is downsampled using the max-pooling operation. The sizes of the kernels in the convolutional and the max-pooling layers are set as $3\times3$ and $2\times2$ respectively. The number of filters in the $4$ convolutional layers ( and $4$ residual blocks) are $32$, $64$, $128$ and $256$ respectively. The encoder takes an image of size $256\times256\times 3$ as the input and produces coarse feature maps of size $16\times16\times256$ as the output.

The decoder takes the coarse feature maps as inputs and performs upsampling and convolution operations to produce dense feature maps to accommodate pixel-wise classification into forged and authentic classes. It has $4$ upsampling layers and $4$ convolutional layers. The upsampling layers upsample the input feature maps and then convolve with trainable filters in the convolutional layers to produce the dense feature maps. In this work, we use an upsampling factor of $2$ at each upsampling layer.  The numbers of filters in the decoder convolutional layers are $32$, $32$, $16$, and $16$ respectively. The kernels in all the convolutional layers are of size $3\times3$. At each convolutional layer, the ReLU activation and the batch normalization are applied. The decoder produces the dense feature maps of size $256\times 256\times32$.
\subsection{Noise-Stream Encoder-Decoder (NSED)}
This stream focuses on learning low-level manipulation-related features. The low-level image features are proven to be helpful in distinguishing different types of image manipulation operations \cite{Bayar_CNN}. These features are also shown to be able to localize different types of forgeries \cite{zhou}. In splicing forgeries, the spliced parts are likely to have noise levels different from those in the authentic regions. In copy-move and content-removal forgeries, the forged regions are generally passed through different types of post-processing operations to make them look visually undetectable. These post-processing operations change the local dependencies of the pixels present in the forged regions and hence introduce low-level artefacts.

The NSED takes the green colour channel of the input image and feeds it to the encoder, which is an FCN, similar to the one in the ISED. The encoder in this stream, however, employs an additional convolutional layer, called the \emph{high-pass filter (HPF) layer} \cite{median_CNN}, \cite{Bayar_CNN}, \cite{zhou}, before the normal convolutional layer. The HPF layer computes the high-pass residuals from the input image to suppress the image contents and enhance the noise contents. In the HPF layer, the kernels follow certain constraints, unlike in a normal convolutional layer, to compute the high-pass residuals. Depending upon the constraints put on the kernels, the HPF layers available in the literature can be categorized as follows: (i) the median filter residual layer, (ii) the constrained convolutional layer, and (iii) the spatial rich model (SRM) filter layer. The median filter residual layer \cite{median_CNN} employs a fixed set of weights in the kernels to compute the median filtered residuals as the output of the HPF layer. The constrained convolutional layer \cite{Bayar_CNN} extracts content-adaptive high-pass residuals by learning weights under a pre-defined constraint, given by 

\begin{equation}\label{constrain}
\begin{aligned}
w_{k}^{1}(0,0)=-1 \\
\text{and} \textbf{} \sum_{l,m\neq 0}w_{k}^{1}(l,m)=1
\end{aligned}
\end{equation}
where, $w_{k}^{1}(l,m)$ denotes the weight at position $(l,m)$ of the $k$th filter and $w_{k}^{1}(0,0)$ denotes the weight at the center of the corresponding filter kernel. The SRM filter layer computes high pass residuals from the input image by applying a fixed set of SRM filters \cite{fridrich2012rich}. The constrained high-pass filters and the SRM filters have recently been shown to be effective in different forensics problems \cite{zhou}, \cite{mantra}. In this work, we have experimented with the constrained convolutional and the SRM filter layers and found that the constrained high-pass filters perform better than the SRM filters. This is also intuitive as the constrained convolutional layer's content-adaptive filters learn the kernels weights from the training data itself. On the other hand, as the SRM filters are fixed filters, they may not be able to extract optimal features for forgery detection tasks. Hence, we have preferred to use the constrained filters for all the experiments reported in this paper.

Therefore, the first layer of the encoder is an HPF layer with $3$ filters of size $5\times 5$. The rest of the encoder has $4$ convolutional layers, $4$ residual blocks and $4$ max-pooling layers. The sizes of the kernels in the normal convolutional and the max-pooling layers are $3\times3$ and $2\times2$ respectively. The numbers of filters in the normal convolutional layers are $32$, $64$, $128$, and $256$. An input of size $256\times256\times1$ is fed to the encoder and the feature map of size $16\times16\times256$ is produced as the output.


The coarse feature maps produced by the encoder are then fed to the decoder, which performs upsampling and convolution operations to produce the dense feature maps. The decoder of this stream has the same architecture as the one in the Image-stream. Therefore, the output of the NSED is the feature maps of size $256\times256\times32$

\subsection{Feature Concatenation and Prediction Layer}
The output dense feature maps of both the streams are concatenated and fed to a single convolutional layer to produce the final prediction. This way of concatenating features from the decoders is known as the \emph{late-fusion} technique. This is because the decoder of each stream processes the coarse features from the corresponding encoder to generate the dense features, which can be thought of as raw decisions about the authenticity of each pixels in the input images.

More specifically, the decoder outputs of both the streams are first concatenated to create the combined feature maps of size $256\times256\times64$. These feature maps are then fed to the final pixel-wise prediction layer, which is a $1\times1$ convolutional layer with sigmoid nonlinearity that produces class probability for each pixel. As there are only two classes in this case, (i.e., manipulated and authentic) this layer produces a single probability map of size equal to that of the input image. The probability map provides the probability of each pixel being classified as either authentic or forged.

\subsection{Learning}
The encoder-decoder network parameters are learned in the training process by minimizing a loss function computed between the ground-truth and the predicted binary masks over a mini-batch of images. There is generally more authentic pixels than the forged ones in a forged image. The classical cross-entropy loss, which is computed as the average over all the pixels, will be more biased towards the authentic classes. This results in a poor performance in classifying the forged pixels while maintaining a high performance in classifying the authentic pixels. To handle this class-imbalance issue, we have experimented with two different loss functions, namely the weighted cross-entropy loss \cite{eigen2015predicting} and the Dice loss \cite{vnet}, given by the following equations:

\begin{equation}\label{CE_loss}
\begin{aligned}
\mathcal{L}_{CE} = -\frac{1}{M} \sum_{c=1}^2 \sum_{i=1}^{M} w_c g_c(i) \log p_c(i)
\end{aligned}
\end{equation}
and
\begin{equation}\label{Dice_loss}
\begin{aligned}
\mathcal{L}_{dice} = 1 - \sum_{c=1}^2 \frac{2\sum_{i=1}^{M} g_{c}(i)p_c(i)}{{\sum_{i=1}^{M} g_c^2(i)} + {\sum_{i=1}^{M}{p_{c}^2(i)}}}
\end{aligned}
\end{equation}
where, $\mathcal{L}_{CE}$ is the weighted cross-entropy loss, $\mathcal{L}_{dice}$ is the Dice loss, $w_c$ is the weighting factor for the class $c$, $g_c(i)$ and $p_c(i)$ are the ground-truth and predicted values for the pixel $i$ being class $c$, and $M$ is the total number of pixels in the batch of images.

The weighted cross-entropy loss handles class-imbalance by assigning a larger weight to the forged pixels than the weights of the authentic ones. We have used the median class weighting \cite{eigen2015predicting}, where the weight of each class is computed as the ratio between the median of all the class frequencies and its own class frequency computed over the entire training dataset. The Dice loss maximizes the overlap between the predicted mask and the ground truth mask for each class, i.e., manipulated and authentic. The value of the Dice loss always lies between $0$ and $1$, irrespective of the number of pixels in each class, and hence solves class-imbalance. We have experimentally found that the Dice loss outperforms the weighted cross-entropy in terms of the generalization accuracy. This is reported in the experimental section. The superior performance of the Dice loss is also reported for different problems, e.g., \cite{deng2018learning}, where the class-imbalance is present. Hence, we propose to use the Dice loss as the preferred loss function for training the proposed network. We also tried the linear combination of the two loss functions, as in \cite{deng2018learning}, but did not observe any improvement over the Dice loss.

The entire network is trained end-to-end on pairs of input images and the corresponding binary masks until it converges. Once the network is trained, we perform the inference on test images for predicting the forged pixels.

\section{Experimental Results}
To show the effectiveness of the proposed method in localizing different types of forgeries, experiments are performed on the following standard forgery datasets: NIST Nimble 2016 (NIST 16) \cite{nist16}, IEEE Forensics Challenge (IFC) \cite{ifs}, Columbia \cite{columbia}, DSO-1 \cite{carvalho_1}, CASIA v1 and CASIA v2 \cite{dong2013casia} datasets. The details about the datasets are as follows:
\begin{itemize}
	\item NIST16 dataset contains $564$ forged images and their corresponding ground-truth binary masks covering the three types of manipulations: splicing, copy-move, and content-removal.
	\item IFC dataset includes $450$ forged images and their corresponding binary masks. The forged images are created using splicing and copy-move operations.
	\item Columbia dataset comprises $180$ spliced images and their corresponding binary masks. 
	\item DSO-1 dataset is a popular splicing dataset and contains $100$ spliced forgeries along with their corresponding ground-truth binary masks.
	\item CASIA v1 and CASIA v2 are two splicing datasets containing $921$ and $5,123$ spliced images, respectively.
\end{itemize} 

The network is implemented in Keras with Tensorflow backend. We have used the \emph{adam} optimizer with a learning rate of $0.00005$.\footnote{The source code will be made publicly available upon acceptance of the manuscript.} The training batch size is fixed to be $16$ pairs of images and corresponding binary masks. 

We have used the following metrics to measure the forgery localization ability of the proposed method: \emph{per-class intersection-over-union (cIoU)} \cite{mag}, the $F1$-score, the \emph{pixel-wise classification accuracy} and the \emph{area under the ROC curve (AUC)}. 

\subsection{Pre-training on Synthetic Dataset}
There is no standard forgery dataset available, which includes sufficiently large numbers of forged images to train deep neural networks. Therefore, we have first pre-trained the proposed network on a synthetic dataset. This dataset contains artificially created forgeries without any post-processing. The forged images present in this dataset try to imitate real forgeries and help the network learn different manipulation-related traces.

In this work, we have used the synthetic dataset created by Bappy \emph{et al.} \cite{bappy} using DRESDEN \cite{dresden}, COCO \cite{lin2014microsoft} and NIST16 datasets. We have used spliced images only from this dataset. For creating the spliced images, the authors have copied objects from COCO dataset and pasted on different authentic images in DRESDEN and NIST16 datasets. From this dataset, we have used $13,470$ spliced images and their corresponding ground-truth binary masks. We have trained the proposed network on this synthetic dataset by using $90\%$ for training and $10\%$ for validation. Once the network converges on this dataset, we save the model for further fine-tuning and testing on different standard forgery datasets.

To see the generalization ability of the pre-trained network, we have checked its performance on NIST16, IFC, Columbia and DSO-1 datasets. The cIoU values on these datasets are presented in Table \ref{cIoU_synthetic}. We have also shown the cIoU values achieved by three recent methods on Columbia and DSO-1 datasets, for comparative analysis. The proposed method achieves the cIoU values of $0.50$, $0.47$, $0.43$ and $0.46$ on NIST16, IFC, Columbia and DSO-1 datasets, respectively. On the other hand, on Columbia and DSO-1 datasets, MFCN \cite{mfcn} achieves the cIoU values of $0.42$ and $0.37$ respectively, ManTra Net \cite{mantra} achieves the cIoU values of $0.58$ and $0.38$ respectively, and MAG \cite{mag} achieves the cIoU values of $0.77$ and $0.56$ respectively. Although the proposed pre-trained network could not outperform MAG, it outperformed MFCN on DSO-1 and Columbia datasets, and ManTra Net on DSO-1 dataset. It is important to note that the results achieved by MFCN, Mantra Net and MAG methods correspond to models trained on realistic forged images, i.e. created manually. On the other hand, the results achieved by the proposed method correspond to model trained on synthetically generated forgeries. These results indicate the ability of the proposed method to learn important forensics features from synthetic forged images that can localize real-life complex forgeries.

\begin{table*}[!htp]
	\centering
	\caption{cIoU values achieved by the proposed network, pre-trained on Synthetic dataset, and two other exsiting methods. '-' denotes the values are not reported.}
	\label{cIoU_synthetic}
	
	\scalebox{1.0}{
		\begin{tabular}{cc|c|c|c|c}
			\toprule
			& \textbf{CASIA v1} & \textbf{NIST16} & \textbf{IFC} & \textbf{Columbia} & \textbf{DSO-1}  \\  \midrule
			MFCN \cite{mfcn}  & -    & -      & -  & 0.42   & 0.37 \\
			ManTra Net \cite{mantra}   & -   & -      & -  & 0.58   & 0.38 \\
			MAG \cite{mantra}   & -   & -      & -  & 0.77   & 0.56 \\
			Proposed (pre-trained) & 0.55  & 0.50           & 0.47  & 0.51   & 0.46  \\ \midrule
		\end{tabular}
	}
	
\end{table*}

\subsection{Fine-tuning and Evaluation on Standard Forgery Datasets}

The pre-trained network is fine-tuned on a training set created from NIST16, IFC and CASIA v2 datasets. We have split NIST16 and IFC datasets into train ($70\%$), validation ($5\%$), and test ($25\%$), as done in \cite{zhou}. We have used all the spliced images of CASIA v2 for training, resulting in a total of $6,093$ images for training. Additionally, we have performed data augmentation by (1) flipping the images both horizontally and vertically, and (2) cropping the images randomly around the manipulated regions to get a zoomed-in version of the images. In this way, we have generated around $40,000$ training images, which help the network learn more diverse manipulation-related features and reduce overfitting. After the model is fine-tuned on these datasets, we have checked the test accuracies on the test images of the above-mentioned datasets using the quantitative measures.

\begin{figure*}[!t]
	\captionsetup[subfigure]{labelformat=empty}
	\setlength{\tempwidth}{.10\linewidth}
	\centering
	\hspace{\baselineskip}
	\columnname{\hspace{0.0cm}Authentic Image}\hfil
	\columnname{\hspace{0.2cm}Forged Image}\hfil
	\columnname{\hspace{0.3cm}Ground-truth}\hfil
	\columnname{\hspace{0.8cm}Prediction}\hfil
	\columnname{\hspace{0.8cm}Overlap}\\

	\subfloat[]{		\rowname{\hspace{4cm}\textbf{Splicing}} \hspace{0.2cm} \fbox{\includegraphics[height = 3cm, width= 2.5cm]{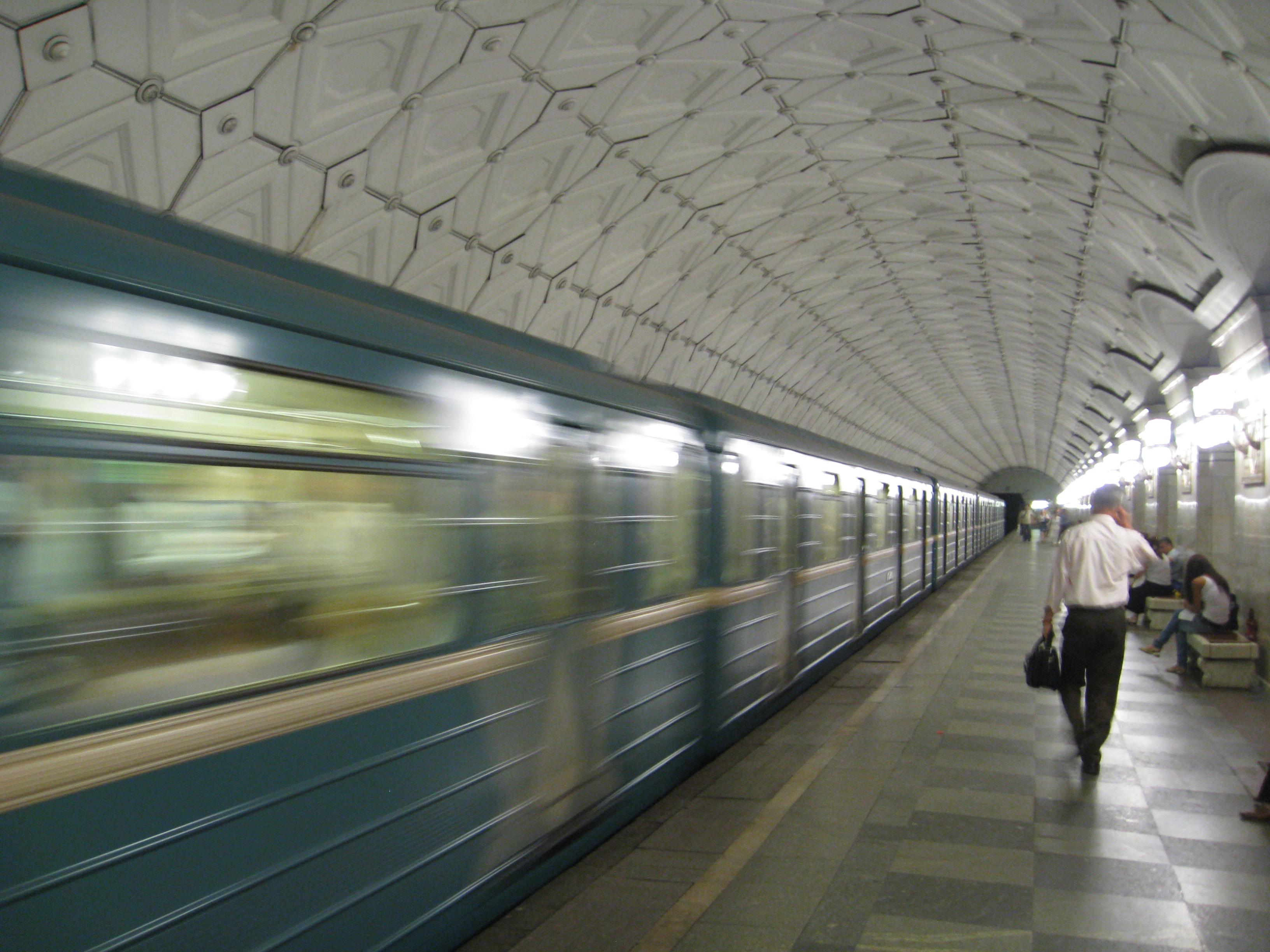}}}\label{a}\hfil
	\subfloat[][]{\fbox{\includegraphics[height = 3cm, width= 2.5cm]{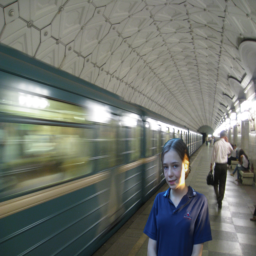}}}\label{b}\hfil
	\subfloat[][]{\fbox{\includegraphics[height = 3cm, width= 2.5cm]{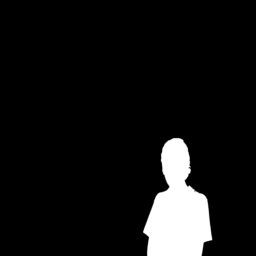}}}\label{b}\hfil
	\subfloat[]{\fbox{\includegraphics[height = 3cm, width= 2.5cm]{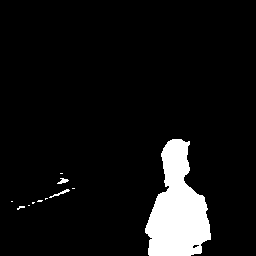}}}\label{c}\hfil
	\subfloat[]{\fbox{\includegraphics[height = 3cm, width= 2.5cm]{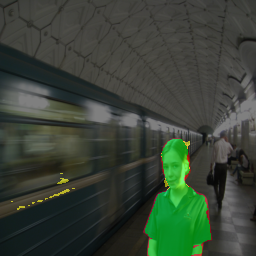}}}\label{d}\\ \vspace{-0.5cm}

	\subfloat[]{\rowname{\hspace{4cm}\textbf{Copy-Move}} \hspace{0.2cm} \fbox{\includegraphics[height = 3cm, width= 2.5cm]{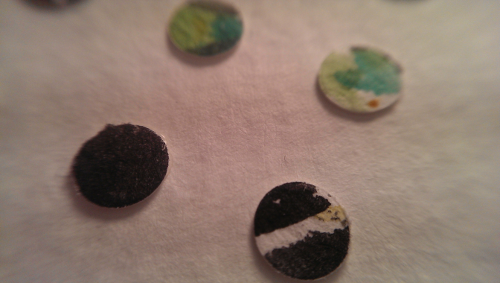}}}\label{d}\hfil
	\subfloat[]{\fbox{\includegraphics[height = 3cm, width= 2.5cm]{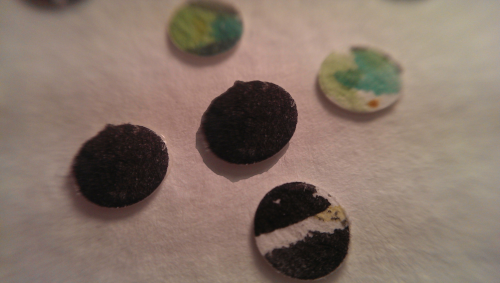}}}\label{e}\hfil
	\subfloat[]{\fbox{\includegraphics[height = 3cm, width= 2.5cm]{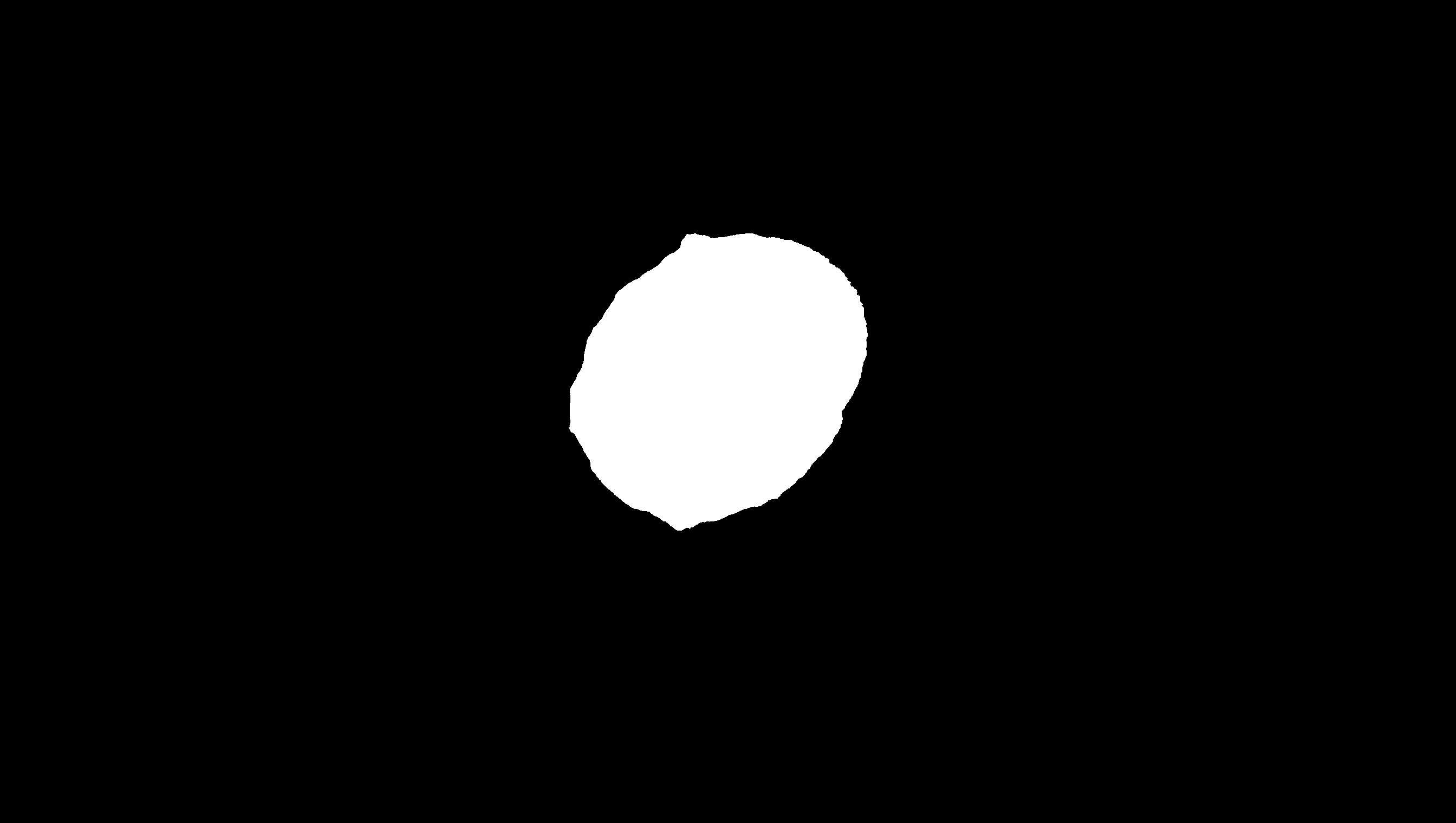}}}\label{e}\hfil
	\subfloat[]{\fbox{\includegraphics[height = 3cm, width= 2.5cm]{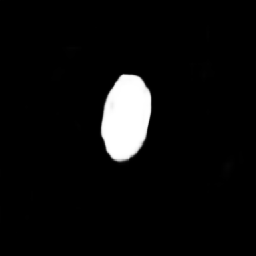}}}\label{f}\hfil
	\subfloat[]{\fbox{\includegraphics[height = 3cm, width= 2.5cm]{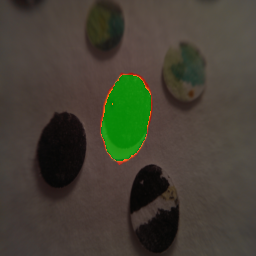}}}\label{g}\\ \vspace{-0.5cm}

	\subfloat[]{\rowname{\hspace{4cm}\textbf{Removal}} \hspace{0.2cm} \fbox{\includegraphics[height = 3cm, width= 2.5cm]{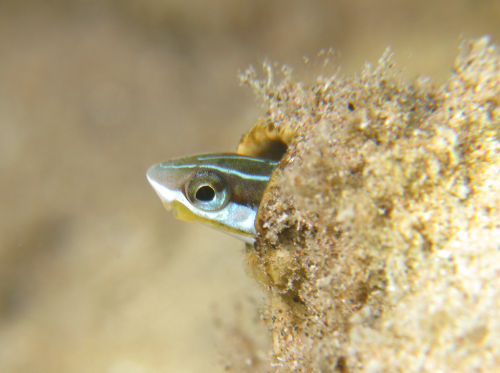}}}\label{d}\hfil
	\subfloat[]{\fbox{\includegraphics[height = 3cm, width= 2.5cm]{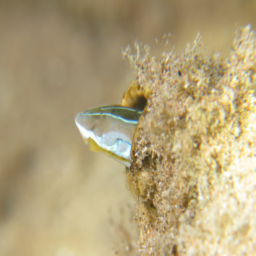}}}\label{e}\hfil
	\subfloat[]{\fbox{\includegraphics[height = 3cm, width= 2.5cm]{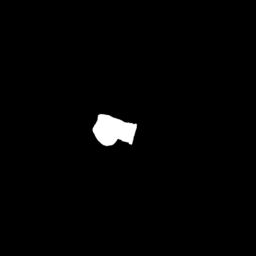}}}\label{e}\hfil
	\subfloat[]{\fbox{\includegraphics[height = 3cm, width= 2.5cm]{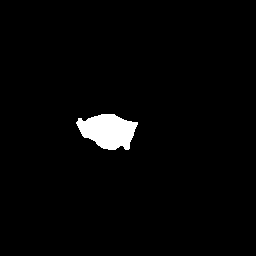}}}\label{f}\hfil
	\subfloat[]{\fbox{\includegraphics[height = 3cm, width= 2.5cm]{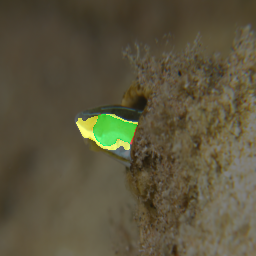}}}\label{g}\\ \vspace{-0.5cm}

	\caption{Forgery localization results of the proposed method for splicing, copy-move and removal forgeries present in NIST16 dataset. First, second, third, fourth, fifth and sixth columns from the left show the authentic image which is used for creating the forgery, the forged image, the ground-truth binary mask, the predicted binary map and the overlap of the ground-truth and binary masks of the proposed method respectively. On the overlap image, the ground-truth, the prediction and overlapped regions are represented by red, yellow and green colours respectively. }
	\label{three_man_loc}
\end{figure*}

\begin{figure*}[!t]
	\captionsetup[subfigure]{labelformat=empty}
	\setlength{\tempwidth}{.10\linewidth}
	\centering
	\hspace{\baselineskip}
	\columnname{\hspace{-1.6cm}Image}\hfil
	\columnname{\hspace{-0.6cm}Ground-truth}\hfil
	\columnname{\hspace{-0.2cm}LSTM-EnDec}\hfil
	\columnname{\hspace{0.6cm}Proposed Method}\\

	\subfloat[]{\fbox{\includegraphics[height = 3cm, width= 2.5cm]{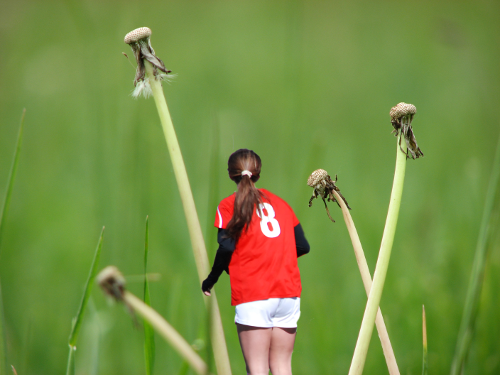}}}\label{d}\hfil
	\subfloat[]{\fbox{\includegraphics[height = 3cm, width= 2.5cm]{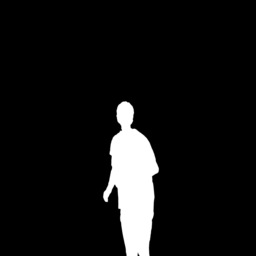}}}\label{e}\hfil
	\subfloat[]{\fbox{\includegraphics[height = 3cm, width= 2.5cm]{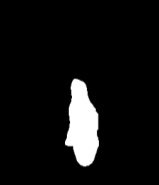}}}\label{e}\hfil
	\subfloat[]{\fbox{\includegraphics[height = 3cm, width= 2.5cm]{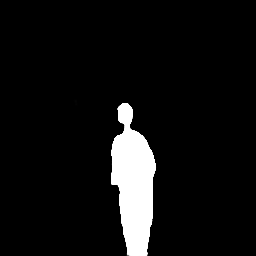}}}\label{f}\\ \vspace{-0.5cm}

	\subfloat[]{\fbox{\includegraphics[height = 3cm, width= 2.5cm]{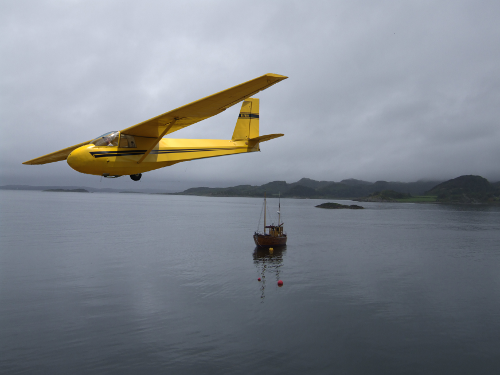}}}\label{d}\hfil
	\subfloat[]{\fbox{\includegraphics[height = 3cm, width= 2.5cm]{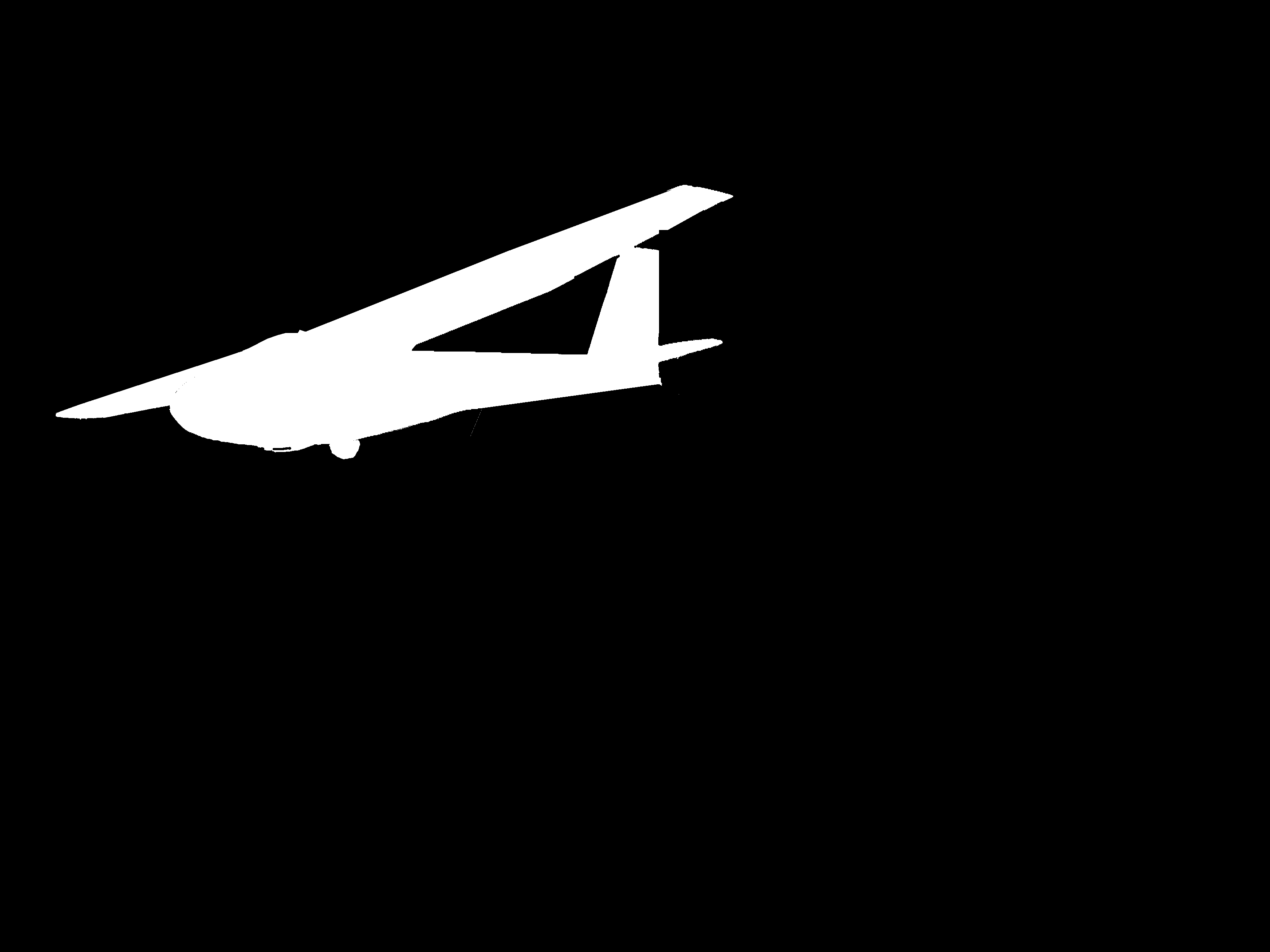}}}\label{e}\hfil
	\subfloat[]{\fbox{\includegraphics[height = 3cm, width= 2.5cm]{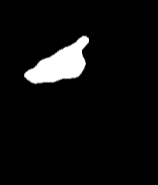}}}\label{e}\hfil
	\subfloat[]{\fbox{\includegraphics[height = 3cm, width= 2.5cm]{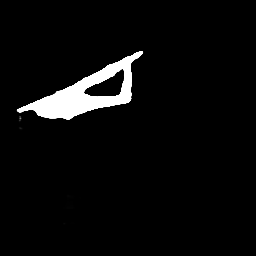}}}\label{f}\\ \vspace{-0.5cm}
	
	\subfloat[]{\fbox{\includegraphics[height = 3cm, width= 2.5cm]{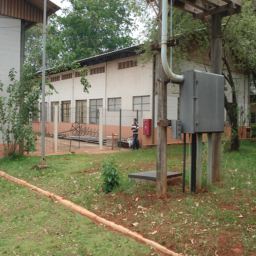}}}\label{a}\hfil
	\subfloat[][]{\fbox{\includegraphics[height = 3cm, width= 2.5cm]{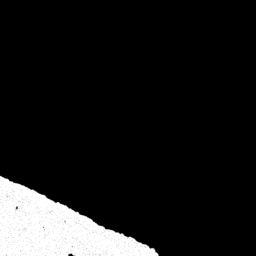}}}\label{b}\hfil
	\subfloat[][]{\fbox{\includegraphics[height = 3cm, width= 2.5cm]{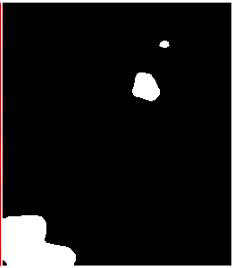}}}\label{b}\hfil
	\subfloat[]{\fbox{\includegraphics[height = 3cm, width= 2.5cm]{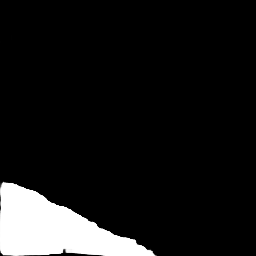}}}\label{c}\\ \vspace{-0.5cm}
	
	\subfloat[]{\fbox{\includegraphics[height = 3cm, width= 2.5cm]{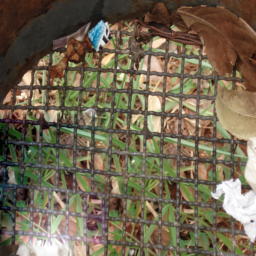}}}\label{a}\hfil
	\subfloat[][]{\fbox{\includegraphics[height = 3cm, width= 2.5cm]{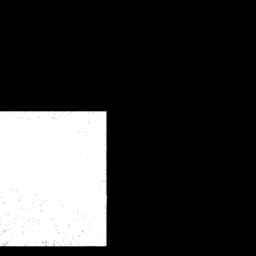}}}\label{b}\hfil
	\subfloat[][]{\fbox{\includegraphics[height = 3cm, width= 2.5cm]{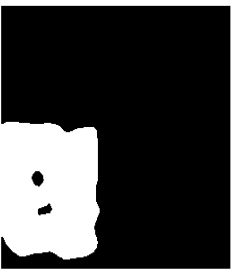}}}\label{b}\hfil
	\subfloat[]{\fbox{\includegraphics[height = 3cm, width= 2.5cm]{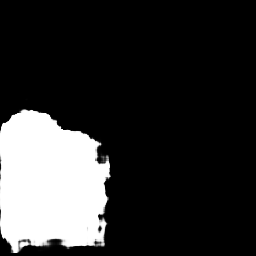}}}\label{c}\\ \vspace{-0.5cm}

	\caption{Examples of qualitative forgery localization results of LSTM-EnDec and the proposed method on NIST16 and IFC datasets. First two rows show the results on images from NIST16 and last two rows show the results on images from IFC dataset. The results of LSTM-EnDec shown in the third column are taken from \cite{bappy}.}
	\label{LSTM_EnDec_Comp}
\end{figure*}

A number of experiments are performed to show the forgery localization ability of the proposed method on various datasets containing different types of forgeries.\par
1) We show the localization ability of the proposed method on the three types of forgeries from NIST16 dataset. Fig. \ref{three_man_loc} shows the localization results on one example image from each manipulation type, i.e., splicing, copy-move and content-removal. We have first computed the pixel-wise accuracies achieved by the proposed method on NIST16 and IFC datasets for quantitative analysis. We have also compared the performance of the proposed method with LSTM-EnDec method, as this method also employs an encoder-decoder network along with an LSTM network. Table \ref{pix_acc} shows the pixel-wise accuracies of the proposed method on these three datasets. It also shows the accuracies achieved by LSTM-EnDec \cite{bappy} on these datasets. The proposed method achieves the pixel-wise accuracies of $95.74\%$ and $92.32\%$ on NIST16 and IFC datasets, respectively.  On the other hand, LSTM-EnDec achieves accuracies of $94.80\%$ and $91.19\%$ on  NIST16 and IFC datasets, respectively. These results quantitatively show the superior performance of the proposed method over LSTM-EnDec on these datasets. Fig. \ref{LSTM_EnDec_Comp} shows some of the qualitative results of LSTM-EnDec and the proposed method on NIST16 and IFC datasets. It can be seen that the proposed method can localize the forged regions better than LSTM-EnDec. The quantitative and qualitative results indicate the ability of the proposed method to learn more discriminative low-level features by employing an encoder network than the hand-crafted features proposed in LSTM-EnDec \cite{bappy}.

\begin{table}[]
	\centering
	\caption{Comparison of the performance of the proposed method with LSTM-EnDec \cite{bappy} on three standard datasets in terms of pixel-wise accuracy.}
	\label{pix_acc}
	\scalebox{1.0}{
		\begin{tabular}{lc|cc}
			\toprule
			& \textbf{NIST16}  & \textbf{IFC}   \\ \midrule
			LSTM-EnDec \cite{bappy}    & 94.80\%              & 91.19\%          \\
			Proposed    & \textbf{95.74}\%            & \textbf{92.32}\%         \\ \midrule
			
		\end{tabular}
	}
\end{table}

\begin{table}[!htp]
	\centering
	\caption{cIoU values on DSO-1 and Columbia datasets.}
	\label{cIoU_all}
	
	\scalebox{1.0}{
		\begin{tabular}{rc|c}
			\toprule
			& \textbf{DSO-1} & \textbf{Columbia}\\  \midrule
			CFA1 \cite{CFA1}        & 0.33           & 0.44              \\ 
			NOI1 \cite{NOI1}        & 0.21           & 0.40              \\ 
			DCT  \cite{DCT}      & 0.24           & 0.41             \\ \midrule
			MFCN \cite{mfcn}      & 0.37           & 0.42              \\
			ManTra Net \cite{mantra}  & 0.38           & 0.58              \\ 
			MAG  \cite{mag}      & \textbf{0.56}  & 0.77              \\ 
			Proposed   & 0.52           & \textbf{0.83}    \\ \midrule
		\end{tabular}
	}
	
\end{table}

\begin{figure}[!htp]
	\centering
	\subfloat[]{\includegraphics[width=9.0cm, height=1.8cm]{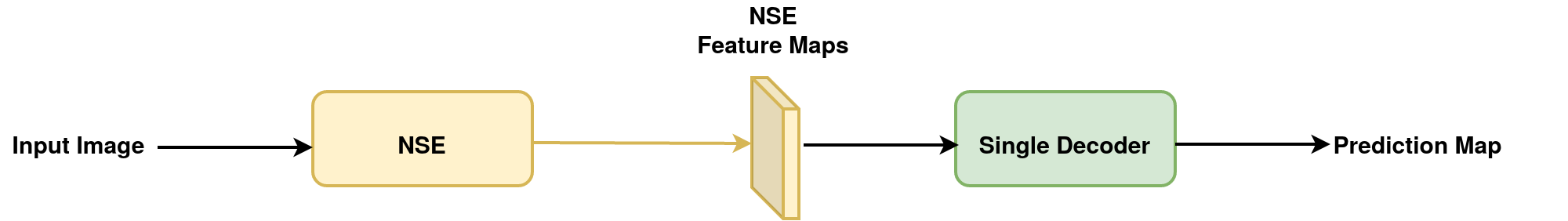} }\\ \subfloat[]{\includegraphics[width=9.0cm, height=3.5cm]{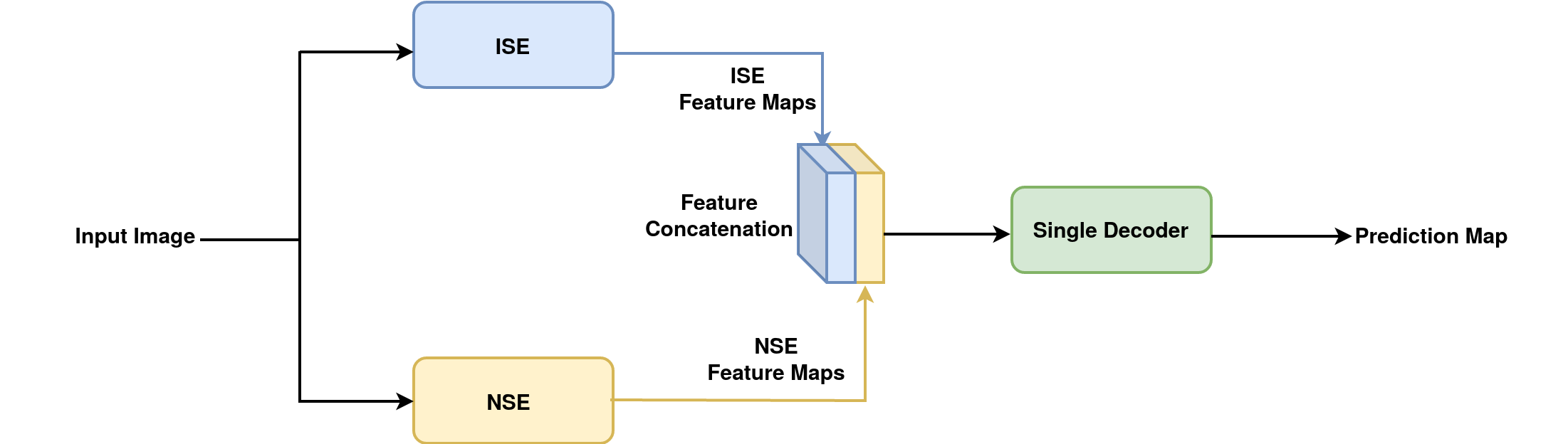} } 
	
	\caption{Different variants of encode-decoder architecture experimented in the paper: (a) NSED and (b) ISE-NSE-1-Dec.} \label{ablation_architecture}
\end{figure}

\begin{table}[!htp]
	\centering
	\caption{F1 scores and AUCs on three datasets.}
	\label{F1_AUC_three_datasets}
	\scalebox{1.0}{
		\begin{tabular}{rccc|ccc|ccc}
			\toprule
			& \multicolumn{3}{c|}{\textbf{NIST16}}                   & \multicolumn{3}{c|}{\textbf{CASIA v1}}                         & \multicolumn{3}{c}{\textbf{Columbia}}                         \\ \midrule
			& \multicolumn{2}{c}{F1}   & AUC             & \multicolumn{2}{l}{F1}            & AUC              & \multicolumn{2}{l}{F1}            & AUC              \\ \midrule
			ELA  \cite{ELA}      & \multicolumn{2}{c}{0.24} & 0.43            & \multicolumn{2}{l}{0.21}          & 0.61             & \multicolumn{2}{l}{0.47}          & 0.58             \\
			NOI1  \cite{NOI1}     & \multicolumn{2}{c}{0.29} & 0.49            & \multicolumn{2}{l}{0.26}          & 0.61             & \multicolumn{2}{l}{0.57}          & 0.55             \\ 
			CFA1   \cite{CFA1}    & \multicolumn{2}{c}{0.17} & 0.50            & \multicolumn{2}{l}{0.21}          & 0.52             & \multicolumn{2}{l}{0.47}          & 0.72             \\ \midrule
			MFCN    \cite{mfcn}   & \multicolumn{2}{c}{0.57} & -               & \multicolumn{2}{l}{\textbf{0.54}}             & -                & \multicolumn{2}{l}{0.57}          & -                \\ 
			RGB-N   \cite{rich_feature}   & \multicolumn{2}{c}{\textbf{0.72}} & 0.94            & \multicolumn{2}{l}{0.41}          & 0.80             & \multicolumn{2}{l}{0.69}          & 0.86             \\ 
			ManTra Net \cite{mantra} & \multicolumn{2}{c}{-}    & 0.80            & \multicolumn{2}{l}{-}             & \textbf{0.82}             & \multicolumn{2}{l}{-}             & 0.82             \\ 
			Proposed   & \multicolumn{2}{c}{0.62} & \textbf{0.95}   & \multicolumn{2}{l}{0.41} & 0.81    & \multicolumn{2}{l}{\textbf{0.86}} & \textbf{0.88}    \\ \midrule
		\end{tabular}
	}
\end{table}

%
%
%

\begin{table}[!htp]
	\centering
	\caption{cIoU values for different compression levels.}
	\label{cIoU_QF}
	
	\scalebox{1.0}{
		\begin{tabular}{rc|c}
			\toprule
			\textbf{Compression Level}	& \textbf{NIST16}  & \textbf{Columbia}\\  \midrule
			
			QF\_50	   & 0.46           & 0.40 \\
			QF\_70	   & 0.47            & 0.44 \\
			QF\_90	   & 0.51           & 0.55 \\
			QF\_100 	   & \textbf{0.72}            & \textbf{0.83}  \\ \midrule
		\end{tabular}
	}
	
\end{table}

2) To show the merits of the proposed method relative to other existing forensics methods, we have compared the performance of the proposed method with the following methods: ELA \cite{ELA}, BLK \cite{BLK}, DCT \cite{DCT}, CFA2 \cite{CFA2}, NOI2 \cite{NOI2}, ADQ2 \cite{ADQ2}, MFCN \cite{mfcn}, RGB-N \cite{zhou}, ManTra Net \cite{mantra} and MAG \cite{mag}.  Table \ref{cIoU_all} shows the cIoU values achieved by the proposed and the competing methods on DSO-1 and Columbia datasets. The cIoU values of the existing methods are taken from \cite{mag}. As can be seen in the table, the proposed method achieves the cIoU values of $0.52$ and $0.83$ on DSO-1 and Columbia, respectively, whereas the best performing method MAG achieves $0.56$ and $0.77$. Although MAG slightly outperforms the proposed method on DSO-1 dataset, it outperforms MAG on Columbia dataset by a large margin.   

\begin{figure*}[!htp]
	\captionsetup[subfigure]{labelformat=empty}
	\setlength{\tempwidth}{.10\linewidth}
	\centering
	\hspace{\baselineskip}
	\columnname{\hspace{-2.6cm}Image}\hspace{0.5cm}
	\columnname{\hspace{-1.0cm}Prediction}\hspace{1.9cm}
	\columnname{\hspace{-0.2cm}Image}\hspace{0.5cm}
	\columnname{\hspace{1.5cm}Prediction}\\
	
	\subfloat[]{\fbox{\includegraphics[height = 3cm, width= 2.5cm]{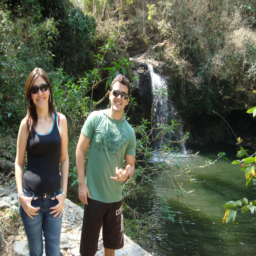}}}\label{a}\hspace{0.5cm}
	\subfloat[][]{\fbox{\includegraphics[height = 3cm, width= 2.5cm]{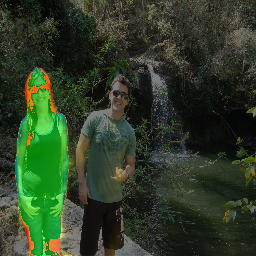}}}\label{b}\hspace{0.5cm} \vrule{} \hspace{0.5cm}
	\subfloat[][]{\fbox{\includegraphics[height = 3cm, width= 2.5cm]{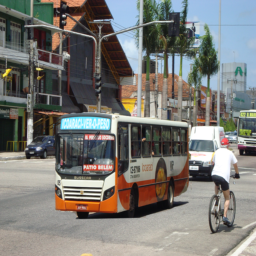}}}\label{b}\hspace{0.5cm}
	\subfloat[]{\fbox{\includegraphics[height = 3cm, width= 2.5cm]{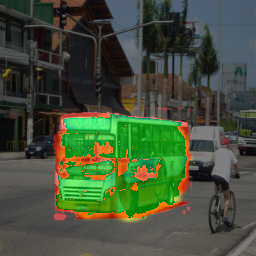}}}\label{c}\\ \vspace{-0.5cm}
	
	\subfloat[]{\fbox{\includegraphics[height = 3cm, width= 2.5cm]{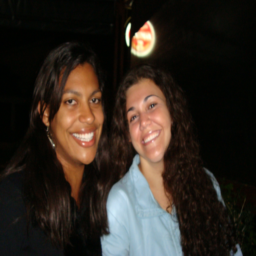}}}\label{a}\hspace{0.5cm}
	\subfloat[][]{\fbox{\includegraphics[height = 3cm, width= 2.5cm]{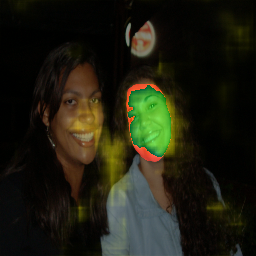}}}\label{b}\hspace{0.5cm} \vrule{} \hspace{0.5cm}
	\subfloat[][]{\fbox{\includegraphics[height = 3cm, width= 2.5cm]{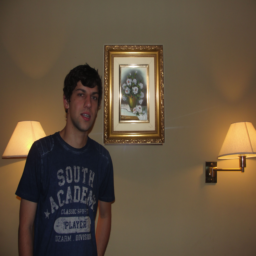}}}\label{b}\hspace{0.5cm}
	\subfloat[]{\fbox{\includegraphics[height = 3cm, width= 2.5cm]{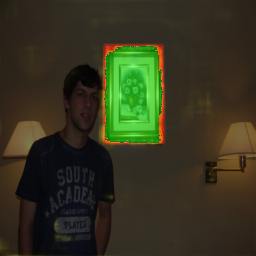}}}\label{c}\\	\rule{0.79\textwidth}{.5pt}
	
	\subfloat[]{\fbox{\includegraphics[height = 3cm, width= 2.5cm]{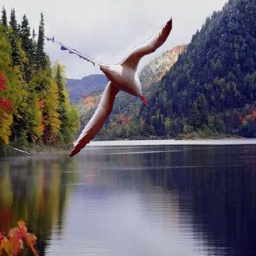}}}\label{a}\hspace{0.5cm}
	\subfloat[][]{\fbox{\includegraphics[height = 3cm, width= 2.5cm]{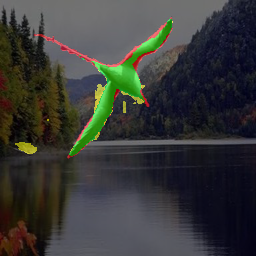}}}\label{b}\hspace{0.5cm} \vrule{} \hspace{0.5cm}
	\subfloat[][]{\fbox{\includegraphics[height = 3cm, width= 2.5cm]{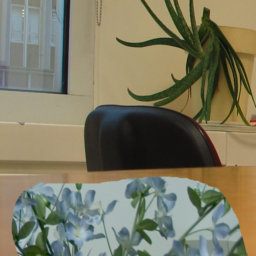}}}\label{b}\hspace{0.5cm}
	\subfloat[]{\fbox{\includegraphics[height = 3cm, width= 2.5cm]{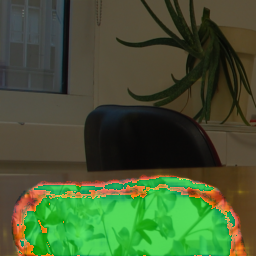}}}\label{c}\\ \vspace{-0.5cm}
	
	\subfloat[]{\fbox{\includegraphics[height = 3cm, width= 2.5cm]{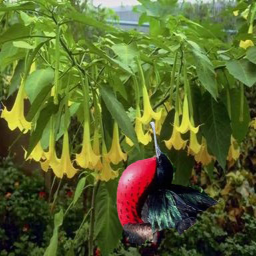}}}\label{a}\hspace{0.5cm}
	\subfloat[][]{\fbox{\includegraphics[height = 3cm, width= 2.5cm]{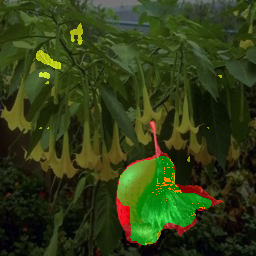}}}\label{b}\hspace{0.5cm} \vrule{} \hspace{0.5cm}
	\subfloat[][]{\fbox{\includegraphics[height = 3cm, width= 2.5cm]{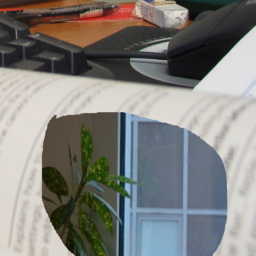}}}\label{b}\hspace{0.5cm}
	\subfloat[]{\fbox{\includegraphics[height = 3cm, width= 2.5cm]{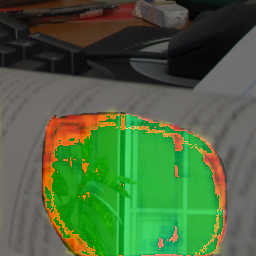}}}\label{c}\\

	\caption{Some qualitative results showing the localization ability of the proposed method on different datasets. The results shown in the top left, the top right, the bottom left and bottom right quadrants are predictions on DSO-1, IFC, CASIA v1 and Columbia datasets respectively.  }
	\label{result_on_four_datasets}
\end{figure*}

Table \ref{F1_AUC_three_datasets} shows the performance of the proposed method in terms of the $F1$-score and the AUC value on three datasets. It also shows the performance of other existing forgery localization methods for comparisons. The $F1$-scores and the AUC values of the existing methods are taken from \cite{zhou} and \cite{mantra}. As shown in the table, the proposed method outperforms all the existing methods on Columbia dataset in terms of both the measures. On NIST16 dataset, the proposed method is outperformed by RGB-N method in terms of the $F1$-score. However, in terms of the AUC value, the proposed method outperforms all the existing methods on NIST16 dataset. These results quantitatively show the superior performance of the proposed method in localizing forgeries over the state-of-the-art. We believe that the superior performance of the proposed network over the state-of-the-art methods is due to the ability to learn both the low-level and the high-level artefacts for pixel-wise forgery localization in a more effective way.
Fig. \ref{result_on_four_datasets} shows two examples of forgery localization from DSO-1, IFC, CASIA v1 and Columbia datasets. These results qualitatively show the ability of the proposed network in localizing different forgeries present in multiple datasets.

3) To see the robustness of the proposed method against JPEG compression, we have compressed the images in NIST16 and Columbia datasets with quality factors $50$, $70$ and $90$. Then, we have checked the performance of the method on these compressed versions of the datasets. Table \ref{cIoU_QF} shows the cIoU values of achieved by the proposed method on these versions. Although the performance of the method degrades as the quality factor reduces, it can still achieves a decent cIoU score of at least $0.4$ at a quality factor as low as 50. This is more than the cIoU values achieved by the non-deep learning methods reported in table \ref{F1_AUC_three_datasets}. The degradation of performance with respect to high JPEG compression (i.e., low quality factor) is expected as most of the low-level image manipulation traces are lost when the image is compressed with a low quality factor.

\subsection{Ablation Study}
We have experimented with different network settings and loss functions to find out the best performing one. Firstly, we have varied the number of encoders and decoders in the proposed method. More specifically, we have exerimented with three network settings: (i) NSED: it is a single noise-stream, encoder-decoder network (shown in Fig. \ref{ablation_architecture}(a)), (ii) ISE-NSE-1-Dec: it employs two-stream encoders, i.e. noise and image-streams, and then the features of both the streams are fused (early-fusion) and a single decoder is employed to compute the prediction (shown in Fig. \ref{ablation_architecture}(b)), and (iii) proposed (two-stream encoder-decoder): it has two parellel encoder-decoder networks, i.e. noise and image-stream encoder-decoder, and performs fusion of the decoder feature maps of both the streams (late-fusion) for producing the prediction. Table \ref{F1_AUC_ablation} shows the performance of the three network variants on NIST16 and Columbia datasets in terms of the $F1$-score and the AUC value. As can be seen, the proposed architecture performs the best on all the three datasets in terms of both measures. These results indicate the necessity of learning both the low-level and high-level features for accurately localizing the forgeries. The results also suggest that the late-fusion technique performs better than early-fusion. 

The possible reason for this is that the features computed by the noise-stream and the image-stream encoders may have different distributions. Hence, in case of early-fusion, a single decoder operating on the concatenated features may not be effective in computing the dense feature maps for accurate predictions. On the other hand, in the late-fusion technique, each stream first computes the dense feature maps individually, which are then concatenated and fed to a final convolutional layer. Hence, the difference in the distributions of the features of the two encoder streams does not affect the performance of the network.

\begin{figure*}[!htp]
	\captionsetup[subfigure]{labelformat=empty}
	\setlength{\tempwidth}{.10\linewidth}
	\centering
	\hspace{\baselineskip}

	\columnname{\hspace{-2.3cm}NSED }\hspace{1.0cm}
	\columnname{\hspace{-1.7cm}ISE-NSE-1-Dec} \hspace{0.7cm}
	\columnname{\hspace{-0.7cm}Proposed}\hspace{0.7cm}
	\columnname{\hspace{2.1cm}W. Cross-Entropy }\hspace{0.5cm}
	\columnname{\hspace{3.4cm}Dice Loss }\\

	\subfloat[]{ \hspace{0.2cm} \fbox{\includegraphics[height = 3.2cm, width= 2.2cm]{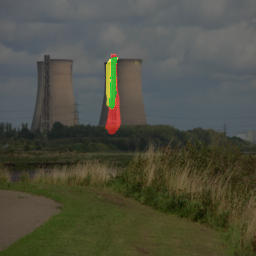}}}\label{a}\hspace{0.7cm}
	\subfloat[][]{\fbox{\includegraphics[height = 3.2cm, width= 2.2cm]{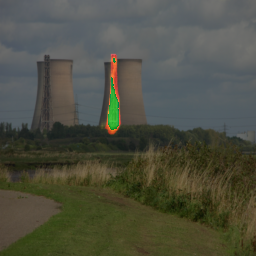}}}\label{b}\hspace{0.7cm}
	\subfloat[]{\fbox{\includegraphics[height = 3.2cm, width= 2.2cm]{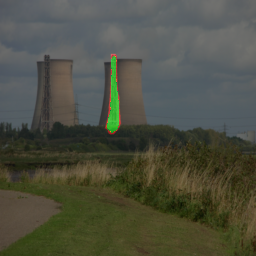}}}\label{d} \hspace{0.7cm}
	\vrule{}\hspace{0.5cm}
	\subfloat[][]{\fbox{\includegraphics[height = 3.2cm, width= 2.2cm]{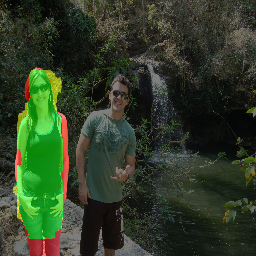}}}\label{b}\hspace{0.5cm}
	\subfloat[]{\fbox{\includegraphics[height = 3.2cm, width= 2.2cm]{splicing-69_2_dec_result.png}}}\label{d} \\ \vspace{-0.5cm}
	
	\subfloat[]{ \hspace{0.2cm} \fbox{\includegraphics[height = 3.2cm, width= 2.2cm]{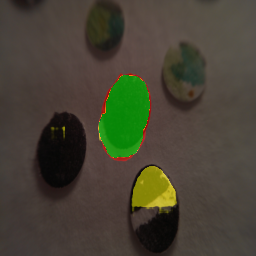}}}\label{a}\hspace{0.7cm}
	\subfloat[][]{\fbox{\includegraphics[height = 3.2cm, width= 2.2cm]{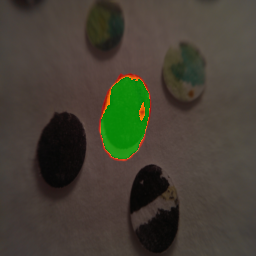}}}\label{b}\hspace{0.7cm}
	\subfloat[]{\fbox{\includegraphics[height = 3.2cm, width= 2.2cm]{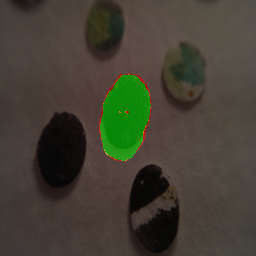}}}\label{d}
	\hspace{0.8cm}\vrule{}\hspace{0.5cm}
	\subfloat[][]{\fbox{\includegraphics[height = 3.2cm, width= 2.2cm]{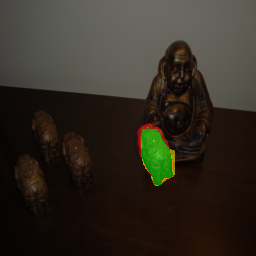}}}\label{b}\hspace{0.5cm}
	\subfloat[]{\fbox{\includegraphics[height = 3.2cm, width= 2.2cm]{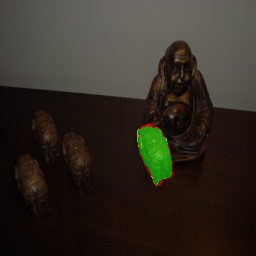}}}\label{d}\\

	\caption{Ablation study of the proposed network, with different network settings and loss functions. Left three images are examples of predictions by NSED, ISE-NSE-1-Dec and the proposed networks with Dice loss respectively on NIST datasets. Last two columns are predictions of the proposed network using weighted cross-entropy and Dice losses resepectively on IFS dataset.}
	\label{dice_vs_ce}
\end{figure*}

Finally, we have checked the performance of the network, when trained with weighted cross-entropy loss instead of Dice loss. Table \ref{cIoU_cost_fn} shows the performance of the proposed method (i.e., two-stream encoder-decoder) on IFC and DSO-1 datasets, when trained using weighted cross-entropy and Dice losses. The results show that on both datasets, the network trained using Dice loss performs better than weighted cross-entropy loss. Fig. \ref{dice_vs_ce} shows the qualitative results of the ablation study.

\begin{table}[!htp]
	\centering
	\caption{Performance comparison of the ablated versions of the proposed network on two datasets.}
	\label{F1_AUC_ablation}
	\scalebox{1.0}{
		\begin{tabular}{cccc|ccc}
			\toprule
			& \multicolumn{3}{c|}{\textbf{NIST16}}                                           & \multicolumn{3}{c}{\textbf{Columbia}}                         \\ \midrule
			& \multicolumn{2}{c}{F1}   & AUC               & \multicolumn{2}{l}{F1}            & AUC              \\ \midrule

			NSED  & \multicolumn{2}{c}{0.51}    & 0.93                    & \multicolumn{2}{l}{0.75}             & 0.85             \\ 
			NSE-ISE-1-Dec  & \multicolumn{2}{c}{0.50}    & 0.92                      & \multicolumn{2}{l}{0.77}             & 0.88             \\ 
			proposed   & \multicolumn{2}{c}{\textbf{0.62}} & \textbf{0.95}     & \multicolumn{2}{l}{\textbf{0.86}} & \textbf{0.88}    \\ \midrule
		\end{tabular}
	}
\end{table}

\begin{table}[!htp]
	\centering
	\caption{Comparison of cIoU values for weighted cross-entropy and Dice losses.}
	\label{cIoU_cost_fn}
	
	\scalebox{1.0}{
		\begin{tabular}{cc|c}
			\toprule
			\textbf{Loss Function}	& \textbf{IFC} & \textbf{DSO-1} \\  \midrule
			
			Weighted Cross-Entropy   & 0.64           & 0.49  \\
			Dice  	   & \textbf{0.68}           & \textbf{0.52}   \\ \midrule
		\end{tabular}
	}
	
\end{table}

\section{Conclusions}
This paper proposed a novel two-stream encoder-decoder network for localizing different types of forgeries, namely splicing, copy-move, and content-removal. One of the streams learns the high-level manipulation-related traces, such as unnatural contrast, from the RGB pixel values in the encoder side. The encoder of the other stream learns the low-level features, such as noise inconsistencies, by employing a high-pass filtering layer as the first layer of the encoder CNN. The decoders of both the streams perform upsampling and convolution on the coarse feature maps computed by the encoders and produce dense feature maps of the same resolution as the input. The dense feature maps of both the streams are concatenated, and fed to a single convolutional layer with sigmoid nonlinearity to produce the pixel-wise probability map. The probability map gives the probability of each pixel being classified as forged or authentic. The experimental results on multiple standard forgery datasets show the effectiveness of the proposed method with respect to the state-of-the-art methods.

\bibliographystyle{IEEEtran}
\bibliography{reference}

\end{document}